\documentclass{mva_style}
\usepackage{graphicx}
\usepackage{enumitem}
\usepackage{seqsplit}
\usepackage{color}
\usepackage{url}
\usepackage{amsmath, amsthm, amssymb, cases, multirow, bm}
\usepackage[pagebackref,breaklinks,colorlinks]{hyperref}
\usepackage[capitalize]{cleveref}
\usepackage[normalem]{ulem}
\useunder{\uline}{\ul}{}

\newcommand{\etal}{\textit{et al}.}
\newcommand{\ie}{\textit{i}.\textit{e}.}
\newcommand{\eg}{\textit{e}.\textit{g}.}

\finalcopy 
\begin{document}

\title{Human Motion Prediction via Test-domain-aware Adaptation with Easily-available Human Motions Estimated from Videos}

\author{
  Katsuki Shimbo ~~~ Hiromu Taketsugu ~~~ Norimichi Ukita\\
  Toyota Technological Institute, Japan\\
  {\tt \{sd25502,ukita\}@toyota-ti.ac.jp}\\
}

\maketitle

\section*{\centering Abstract}
\textit{
  In 3D Human Motion Prediction (HMP), conventional methods train HMP models with expensive motion capture data. However, the data collection cost of such motion capture data limits the data diversity, which leads to poor generalizability to unseen motions or subjects.
  To address this issue, this paper proposes to enhance HMP with additional learning using estimated poses from easily available videos. The 2D poses estimated from the monocular videos are carefully transformed into motion capture-style 3D motions through our pipeline. By additional learning with the obtained motions, the HMP model is adapted to the test domain. The experimental results demonstrate the quantitative and qualitative impact of our method.
}

\section{Introduction}
Human Motion Prediction (HMP) is the problem of forecasting human motion, \ie, the sequence of human poses, each of which is represented by the 3D coordinates of human body joints. 
By incorporating it into robots, we can develop collaborative robots~\cite{collabo} and guide robots~\cite{guide} that behave proactively.

In most cases, training data for 3D HMP has been collected by motion capture systems~\cite{gpmgm,FirstRNN,rnn2,LTD,HumanMAC}.
However, motion capture measurement requires significant effort and environmental constraints, limiting data collection to only basic actions, such as walking, performed by a small number of subjects. HMP models trained on such limited data exhibit poor performance for unseen test motions and subjects.

\begin{figure}[t]
\centering
\includegraphics[width=\linewidth]{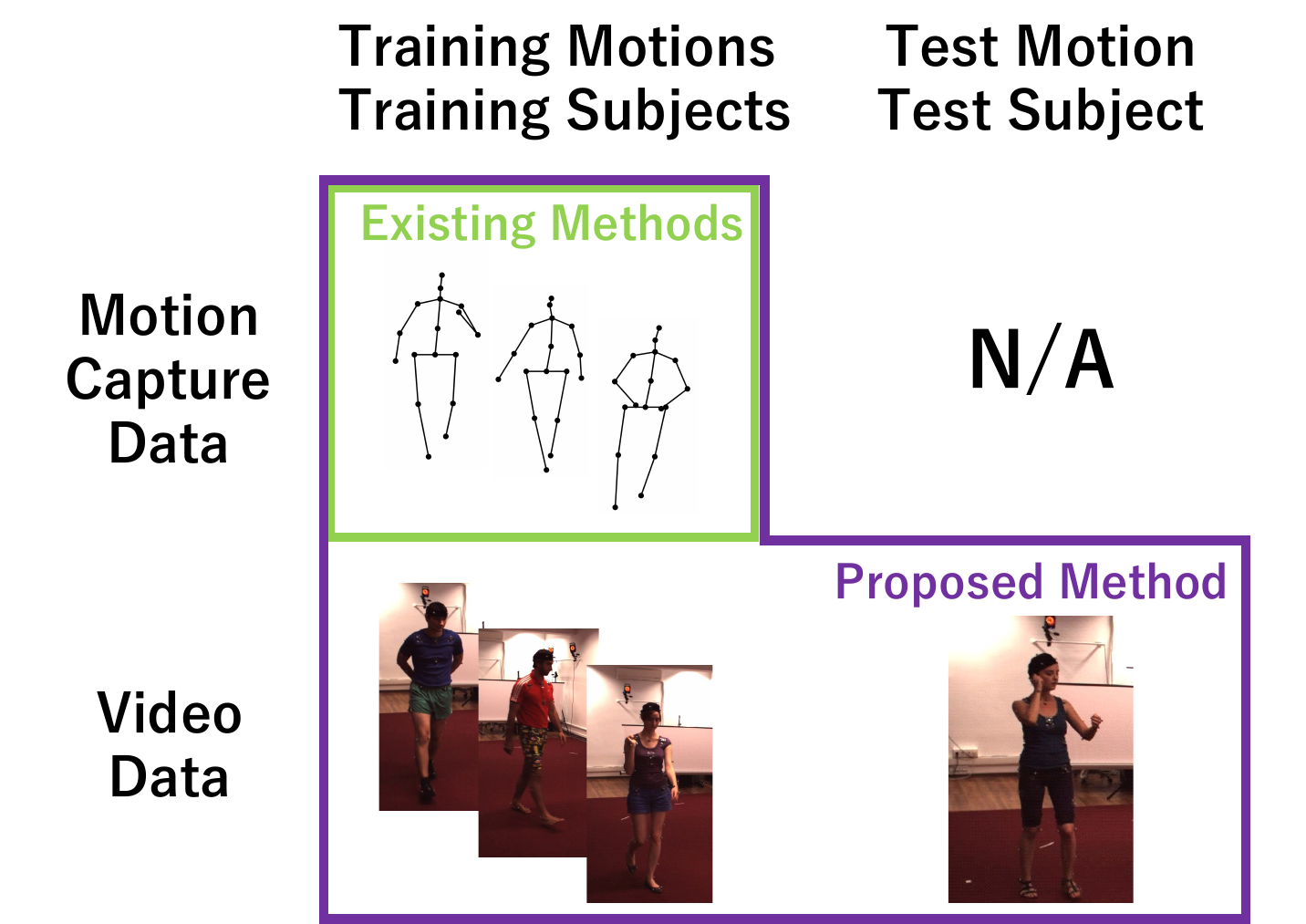}
  \caption{Comparison of data usage in the proposed and existing methods. In addition to the motion capture data, test motions (NOT test data) estimated from videos of a test subject are used for adaptation in the proposed method.}
  \label{fig:1}
\end{figure}

To increase the amount of training data~\cite{personalization,dozens}, this paper proposes a method that utilizes easily obtainable videos captured in a domain similar to the test data. While conventional methods train only with motion capture data of other people, our approach additionally utilizes 3D human motions estimated from the videos of test subjects’ motions, as illustrated in Fig.~\ref{fig:1}.
Note that although additional training videos share the same domain as the test videos, these videos are different from the test videos. Instead, after a small number of test subjects' videos are collected and used for training the motion predictor for personalization, other new videos of these test subjects are given to the motion predictor at inference.

For the additional training, 3D human poses estimated from videos must be as accurate as possible. To this end, we employ the 3D human pose estimator pre-trained with large-scale data~\cite{motionbert}. Furthermore, to enhance motion prediction performance by incorporating 3D human poses estimated from videos, the joint positions of the estimated poses must align with those of motion capture data, which are used to train the motion predictor.
However, conventional 3D pose estimators~\cite{2dposecontext,poseformer,poseformerv2} are trained with their training datasets, in which the joint positions are defined differently from those of the motion predictor.
To address this issue, our method introduces a human mesh model~\cite{SMPL} as an intermediary. From this mesh data, our method extracts the joints so that they are defined as the same structure as the motion predictor.

Our contributions are as follows:
\begin{enumerate}
\item We propose additional learning with 3D motions estimated from test-domain videos for HMP.
\item Our pipeline aligns the estimated poses with motion capture data using a human mesh model.
\item Our method quantitatively and qualitatively enhances the HMP performance against the baseline.
\end{enumerate}

\section{Related Work}
While many trajectory prediction methods~\cite{flowchain,physics} are proposed and highly matured, full-body HMP is much difficult due to its high dimensionality and stochasticity. To make matters worse, the data collection cost is much higher, leading data shortage.

\paragraph{Data augmentation in HMP.}
To mitigate the data shortage, MotionAug~\cite{MotionAug} augmented HMP data using Variational Auto-Encoder~\cite{VAE}. However, since the amount of real data is extremely limited, this method alone cannot adequately handle the unseen test subjects’ motions. Our method addresses this issue by incorporating additional videos during testing.

\paragraph{Test-domain adaptation in HMP.}
H/P-TTP~\cite{personalization} adapts a motion predictor to test subjects by leveraging past motion data used as input to the motion prediction network during testing. However, our study offers two advantages over this method: a more practical experimental setup and reduced time cost.

First, regarding the experimental setup,  H/P-TTP assumes that accurate motion capture data is available when training the motion predictor at inference. However, in real-world scenarios, this assumption does not hold. In contrast, we assume that test subjects’ motions are recorded as videos before deployment, making the experimental setup more realistic and feasible.

Next, regarding time cost,  H/P-TTP requires retraining the motion predictor each time the past motions are input during deployment. This can be a significant challenge for HMP which requires real-time processing due to the high computational cost of training. In contrast, our method learns from pre-recorded videos of the test subjects' motions before deployment, eliminating the need for training during operation.

\section{Proposed Method}
\label{section:Proposedmethod}
\begin{figure}[t]
  \begin{center}
     \includegraphics[width=\columnwidth]{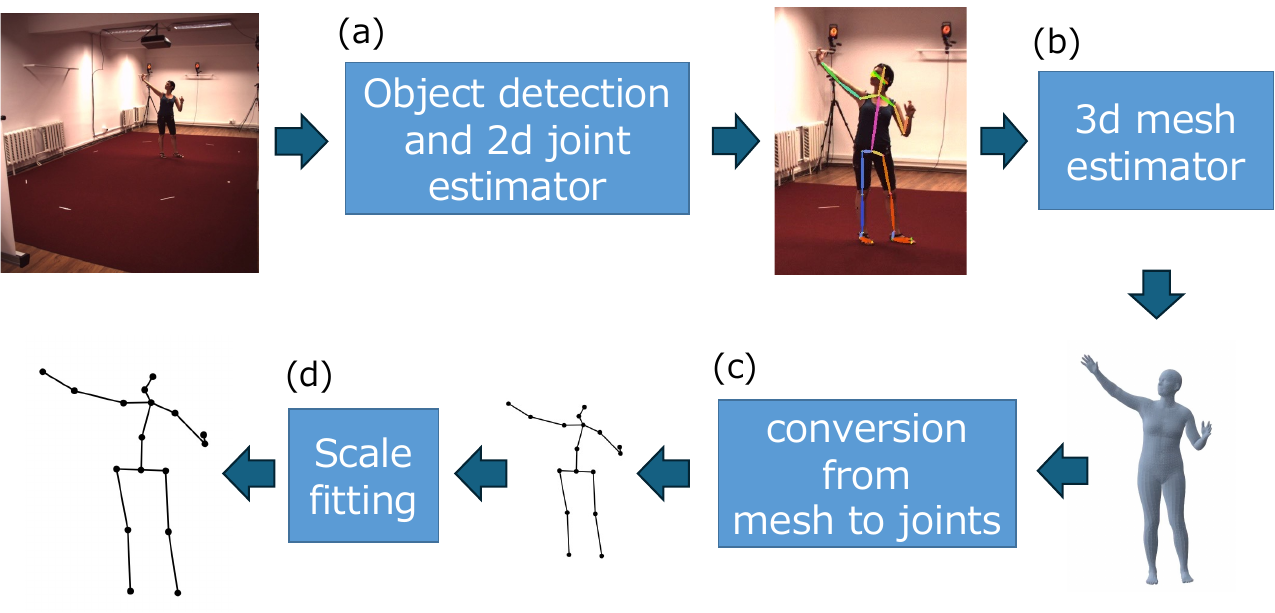}
     \caption{Method for generating 3D poses from videos. Motion sequences are obtained by processing each frame of the video.} 
     \vspace{-3mm}
     \label{fig:motionpred}
   \end{center}
\end{figure}

\vspace{-1mm}
\paragraph{Overview.}
Figure~\ref{fig:motionpred} illustrates the process of generating 3D poses from test-domain videos. First, (a) 2D human joints are extracted from the test-domain videos, and (b) the corresponding 3D mesh is estimated (Sec.~\ref{subsection:estimate3dpose}). Next, using the 3D mesh as an intermediate representation, (c) the joint positions are converted to match the pose definition of the motion capture data, and (d) the scale is adjusted to ensure temporal consistency across the sequence (Sec.~\ref{subsection:ScaleAdjustment}). Finally, the resulting 3D poses are added to the training data to train the human motion predictor (Sec.~\ref{subsection:training}).

\subsection{3D Mesh Estimation as Intermediary}
\label{subsection:estimate3dpose}
\vspace{-1mm}
\paragraph{Object detection and 2D joint estimation.}
To estimate a 3D pose, we first detect the human regions in the videos by object detection and input the detected regions into a 2D pose estimator to estimate the 2D poses in each frame, as shown in Fig.~\ref{fig:motionpred} (a).

\paragraph{3D mesh estimation.}
In general, after obtaining the 2D pose, the next step is to use a 3D pose estimator to estimate the 3D pose. However, training a 3D pose estimator requires paired 2D and 3D pose data, assuming that the number and definitions of joints are consistent. If the joint definitions used in the 2D and 3D pose data for training the 3D pose estimator differ from those in the motion capture data, the estimated 3D poses cannot be used for motion prediction.

For example, if we aim to predict the kicking motion in soccer, joints such as the toes and ankles are crucial. However, if the 3D pose estimator does not estimate these joints, the predicted poses become unsuitable for motion prediction. To address this issue, our study first estimates a 3D human mesh as shown in Fig.~\ref{fig:motionpred} (b). This enables us to adjust joint positions based on the target motion, and by utilizing a human mesh representation as an intermediary, we can create poses that do not rely on specific joint definitions of motion capture data or 2D pose estimation methods.  

\begin{figure*}[t]
  \begin{center}
     \includegraphics[width=0.85\textwidth]{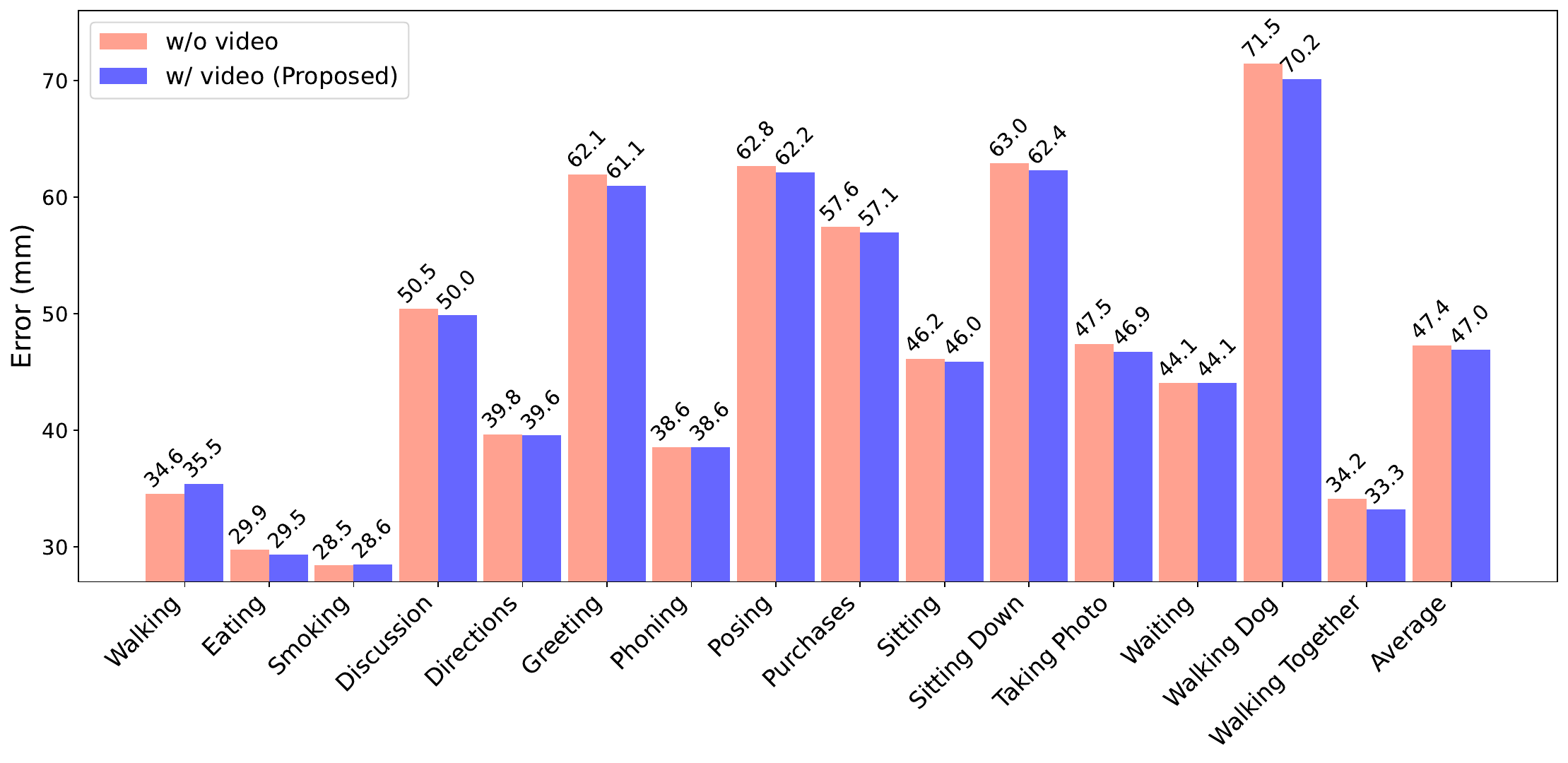}
     \caption{Comparison of the prediction error (mm) between using (blue, ours) and not using (pink, baseline) videos of the test action at 400 msec. } 
     \label{fig:results}
   \end{center}
   \vspace{-10mm}
\end{figure*}

\subsection{3D Pose Conversion from Mesh}
\label{subsection:ScaleAdjustment}
\paragraph{Conversion from mesh to joints.}
The 3D human mesh model has its unique joint definitions, where the skeleton is programmed to determine how the surface shape moves based on joint movements. When estimating a 3D mesh from a 2D pose, it is necessary not only to align the approximate joint definitions between them but also to refine the surface shape using the 3D human mesh silhouette. This process ultimately allows for refining the estimated 3D pose. Finally, by converting the surface shape of the mesh into the joint definitions used in motion capture datasets, we obtain a 3D motion that matches the joint count and definitions of motion capture data, as shown in Fig.~\ref{fig:motionpred} (c).

\paragraph{Scale fitting.}
Although the aforementioned process allows us to convert mesh to 3D poses, poses estimated from videos are still noisy, \eg, the estimated human scales often vary across frames. Therefore, it is necessary to align the scale of the 3D poses obtained from video data with that of motion capture data. In this study, we consider the consistency of inter-joint distances over time as an essential condition for ensuring accurate human pose, similar to motion capture measurements. Our proposed method adjusts the scale of each joint-to-joint vector on a per-frame basis to match the inter-joint lengths defined in the motion capture data, as shown in Fig.~\ref{fig:motionpred} (d). 

Specifically, the following steps are performed:
\begin{enumerate}[label=\textbf{Step\arabic*.}, leftmargin=13mm]
    \item Compute the 3D directional vector connecting a central joint (\eg, right shoulder) and an adjacent joint extending towards the extremity (\eg, right elbow) from the relative 3D pose obtained from the video data.
    \item Normalize this directional vector to obtain a 3D unit vector and replace its magnitude with the static offset value defined in the skeletal structure. This ensures that the inter-joint distances are adjusted to match those in the motion capture dataset.
    \item Repeat the above process from the central joint to the end joints. For example, after adjusting the right shoulder to the right elbow, proceed sequentially from the right elbow to the right wrist. Apply the same process to other limbs (\eg, left arm, legs .., etc).
\end{enumerate}

\subsection{Training the Motion Predictor}
\label{subsection:training}
\paragraph{Test-domain-aware adaptation.}
We assume a scenario where one of the actions in the dataset is considered a novel test motion, and the motion capture data for test motion is unavailable for training.
Instead, this setup assumes that video data of the test motions performed by the test subject is available.

This assumption is justified when considering applications like collaborative or household robots, where the system is intended for use by a particular individual.
It enables additional learning from motions in test-domain videos without annotations.

\paragraph{Loss function.}
In our method, we use Mean Per Joint Position Error (MPJPE)~\cite{h36m} as the training loss as well as the existing HMP methods~\cite{rnn2,LTD}.
The MPJPE at frame $n$ is defined as follows:
\begin{eqnarray}
{\bf MPJPE}(n) = \frac{1}{J} \sum_{j=1}^{J} \|\hat{\textbf{p}}_{j,n} - \textbf{p}_{j,n}\|^{2},
\end{eqnarray}
where $\hat{\textbf{p}}_{j,n} \in \mathbb{R}^{3}$ represents the predicted position of the $j$-th joint at frame $n$, and $\textbf{p}_{j,n} \in \mathbb{R}^{3}$ represents the corresponding ground truth position.

Since HMP involves forecasting over $N$ frames, the total loss function is computed as follows:
\begin{eqnarray}
\mathcal{L} = \frac{1}{J \times N} \sum_{n=1}^{N} \sum_{j=1}^{J} \|\hat{\textbf{p}}_{j,n} - \textbf{p}_{j,n}\|^{2}.
\end{eqnarray}

\begin{table}[t]
 \caption{Average errors of 15 actions. ``w/ Ground Truth'' denotes the additional training with motion capture data of the test-domain videos. The smaller errors are bolded.}
 \label{table:comp}
 \begin{center}
 \vspace{-3mm}
 \scalebox{0.9}[0.9]
   {
    \begin{tabular}{c|cccc}
    \hline
                                & \multicolumn{4}{c}{Average}                                                                                  \\ \hline
    (msec)                      & 80                       & 160                       & 320                       & 400                        \\
    w/o video (Baseline)      & \textbf{9.18}                     & 19.01                     & 38.63                     & 47.30                      \\
    w/ video (Ours)   & \textbf{9.18}                     & \textbf{18.92}                     & \textbf{38.29}                     & \textbf{46.91} \\
    \multicolumn{1}{l|}{w/ Ground Truth} & \multicolumn{1}{l}{8.99} & \multicolumn{1}{l}{18.54} & \multicolumn{1}{l}{37.62} & \multicolumn{1}{l}{46.17} \\ \hline
    \end{tabular}}
    \vspace{-5mm}
    \end{center}
    \end{table}

\begin{figure}[t!]
  \begin{center}
  \includegraphics[width=0.9\columnwidth]{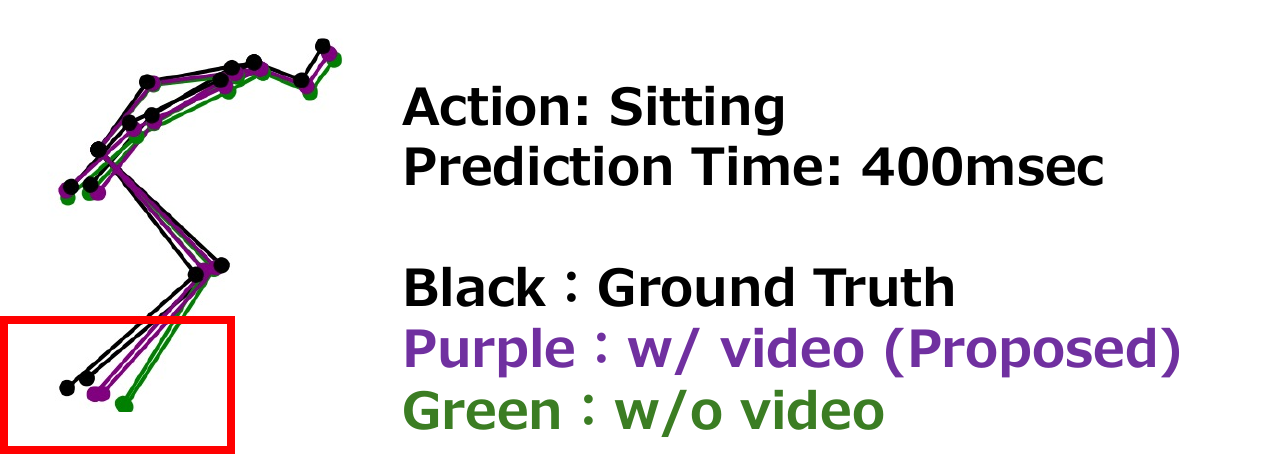}
     \caption{Comparison between the proposed method (purple) and the baseline (green).}
     \label{fig:h36m_quali}
     \vspace{-8mm}
   \end{center}
\end{figure}

\section{Experiments}
\vspace{-2mm}

\subsection{Experimental Setup}
\label{subsection:setup}
\vspace{-2mm}
\paragraph{Datasets.}
\label{paragraph:datasets}
We selected Human3.6M~\cite{h36m} as it contains both videos and motion capture data. It consists of 15 actions performed by 7 subjects, recorded by four cameras. Actions include walking, eating, and so on.

For each action, subject, and camera, there exist two videos.
Therefore, one of the recordings is used as the video for additional training, while the other is used for evaluation. There are two choices regarding which recording to use for training and testing.
We present their average values based on both choices.

Each action is recorded with high-quality 3D joint positions by tracking markers attached to the subject’s body using the Vicon motion capture system~\cite{vicon}.
To align the experimental conditions with previous studies~\cite{rnn2,LTD}, we remove global rotation and global translation. For the same reason, both video and motion capture sequences are downsampled to 25 FPS.

The subjects used for training are 1, 6, 7, 8, and 9; 11 and 5 are used for validation and testing, respectively.
\vspace{-3mm}
\paragraph{Baselines.}
\label{paragraph:baselines}
For 2D pose estimation in Fig.~\ref{fig:motionpred}(a), we adopt Alphapose~\cite{AlphaPose}. The original implementation of Alphapose~\cite{AlphaPose} processes four images at a time, and if only one image fails to detect a person, no pose data can be obtained. To maximize the available poses, we process images one by one, utilizing video data from all four directions. For 3D human mesh estimation, we employ MotionBERT~\cite{motionbert}, which is pre-trained with large-scale data. As the human motion predictor, we employ LearnTrajDep, the representative approach, proposed by Mao~\etal~\cite{LTD}. Our method maintains inference cost with the baseline; the inference time for HMP to output 400 msec of future motions is 55 msec with the NVIDIA RTX A6000 GPU.

For the training of baselines, we followed the settings in the respective papers or implementations.
\vspace{-3mm}
\paragraph{Metrics.}
\label{paragraph:Evaluation Metrics}
We evaluate MPJPE, which measures the prediction error as described in Sec.~\ref{subsection:training}, for the 16 joints used in previous studies~\cite{HumanMAC,BeLFusion} as well.
Given 10 frames (400 msec) of past motion, the motion predictor predicts 10 frames of future motion. The evaluation is conducted at 2nd (80 msec), 4th (160 msec), 8th (320 msec), and 10th (400 msec) frames of the output.

\begin{table}[t]
  \caption{Average prediction error (mm) in long-term prediction. The smaller errors are bolded.}
  \label{table:mix2}
  \begin{center}
  \vspace{-7mm}
   \scalebox{0.6}[0.6]
   {
    \begin{tabular}{c|cc|cc|cc|cc}
   \hline
          & \multicolumn{2}{c|}{Walking}     & \multicolumn{2}{c|}{Eating}      & \multicolumn{2}{c|}{Smoking}          & \multicolumn{2}{c}{Discussion}   \\ \hline
(msec)    & 560            & 1000            & 560            & 1000            & 560               & 1000              & 560             & 1000            \\
w/o video & 44.05          & \textbf{48.35}  & 39.85          & 56.70           & \textbf{38.50}    & 54.80             & \textbf{66.80}  & \textbf{90.30}  \\
w/ video  & \textbf{43.40} & 48.50           & \textbf{39.00} & \textbf{56.45}  & \textbf{38.50}    & \textbf{53.90}    & 68.15           & 91.90           \\ \hline
          & \multicolumn{2}{c|}{Directions}  & \multicolumn{2}{c|}{Greeting}    & \multicolumn{2}{c|}{Phoning}          & \multicolumn{2}{c}{Posing}       \\ \hline
(msec)    & 560            & 1000            & 560            & 1000            & 560               & 1000              & 560             & 1000            \\
w/o video & 54.20          & \textbf{76.30}  & 79.05          & 101.75          & \textbf{52.10}    & 77.65             & 89.35           & \textbf{136.15} \\
w/ video  & \textbf{54.15} & 76.80           & \textbf{78.60} & \textbf{101.15} & \textbf{52.10}    & \textbf{77.40}    & \textbf{88.40}  & 136.20          \\ \hline
          & \multicolumn{2}{c|}{Purchases}   & \multicolumn{2}{c|}{Sitting}     & \multicolumn{2}{c|}{Sitting Down}     & \multicolumn{2}{c}{Taking Photo} \\ \hline
(msec)    & 560            & 1000            & 560            & 1000            & 560               & 1000              & 560             & 1000            \\
w/o video & \textbf{76.10} & 109.95          & 64.20          & 99.55           & 85.10             & 126.40            & \textbf{67.05}  & 104.55          \\
w/ video  & 76.75          & \textbf{109.55} & \textbf{63.75} & \textbf{98.65}  & \textbf{84.65}    & \textbf{126.25}   & 67.75           & \textbf{104.25} \\ \hline
          & \multicolumn{2}{c|}{Waiting}     & \multicolumn{2}{c|}{Walking Dog} & \multicolumn{2}{c|}{Walking Together} & \multicolumn{2}{c}{Average}      \\ \hline
(msec)    & 560            & 1000            & 560            & 1000            & 560               & 1000              & 560             & 1000            \\
w/o video & 60.45          & \textbf{83.55}  & 86.00          & 115.85          & 43.30             & 50.20             & 63.06           & 88.80           \\
w/ video  & \textbf{59.95} & \textbf{83.55}  & \textbf{84.35} & \textbf{114.95} & \textbf{42.75}    & \textbf{48.95}    & \textbf{62.82}  & \textbf{88.56}  \\ \hline
    \end{tabular}
    }
    \vspace{-6mm}
\end{center}
\end{table}

\subsection{Baseline Comparison in Short-term HMP}
\label{subsection:results_short}
In Fig.~\ref{fig:results}, We present a comparison of the results for each action at a 400 msec prediction horizon. 
``w/ video'' denotes that the training data consists of ``the motion capture data of the remaining 14 actions'' and ``motions estimated from the video data of the test action of the test subject''. When video data is not used, the condition is referred to as ``w/o video''.
For actions like {\bf Walking Dog}, which involves complex movements such as being pulled by a dog, the prediction accuracy showed a notable improvement.
In contrast, for simple actions like {\bf Walking}, the prediction accuracy deteriorated due to the limited advantage of using videos.

One of the qualitative results is shown in Fig.~\ref{fig:h36m_quali}. This illustrates that learning motion patterns from test-domain videos allows for more accurate predictions that are closer to the ground truth.

Next, Table \ref{table:comp} presents a comparison of the results obtained from three different settings: training without video data (w/o video), with video data (w/ video), and using the corresponding motion capture data instead (w/ Ground Truth). While video data has not yet fully replaced motion capture data, advancements in 2D pose estimation and 3D mesh estimation are expected to further bridge this gap.

\subsection{Baseline Comparison in Long-term HMP}
\label{subsection:results_long}
Table~\ref{table:mix2} shows that the proposed method can reduce errors even for long‐term HMP at 560 and 1000 msec. This finding is reasonable since compared to short‐term motions dominated by physical inertia, differences among actions and subjects are expected to become larger at the longer prediction horizon.

\section{Conclusion}
\label{section:conclusion}
This study proposed a method that adopts HMP models to the test domain with easily available videos.
As a result, additional learning with a single video from the test domain reduced both short- and long-term prediction errors for most actions.
Future work includes more efficient test-time adaptation from videos~\cite{stream} and additional learning at scale.

\bibliographystyle{unsrt}
\raggedright
\bibliography{main}

\begin{thebibliography}{10}

\bibitem{collabo}
Mastrogiacomo~Luca Gervasi~Riccardo and Franceschini Fiorenzo.
\newblock A conceptual framework to evaluate human-robot collaboration.
\newblock {\em Int. J. Adv. Manuf. Technol.}, 108(3), 2020.

\bibitem{guide}
Mohammad Mahdavian, Payam Nikdel, Mahdi Taherahmadi, and Mo~Chen.
\newblock {STPOTR:} simultaneous human trajectory and pose prediction using a non-autoregressive transformer for robot follow-ahead.
\newblock In {\em ICRA}, 2023.

\bibitem{gpmgm}
Norimichi Ukita and Takeo Kanade.
\newblock Gaussian process motion graph models for smooth transitions among multiple actions.
\newblock {\em Comput. Vis. Image Underst.}, 116(4), 2012.

\bibitem{FirstRNN}
Katerina Fragkiadaki, Sergey Levine, Panna Felsen, and Jitendra Malik.
\newblock Recurrent network models for human dynamics.
\newblock In {\em ICCV}, 2015.

\bibitem{rnn2}
Julieta Martinez, Michael~J. Black, and Javier Romero.
\newblock On human motion prediction using recurrent neural networks.
\newblock In {\em CVPR}, 2017.

\bibitem{LTD}
Wei Mao, Miaomiao Liu, Mathieu Salzmann, and Hongdong Li.
\newblock Learning trajectory dependencies for human motion prediction.
\newblock In {\em ICCV}, 2019.

\bibitem{HumanMAC}
Ling{-}Hao Chen, Jiawei Zhang, Yewen Li, Yiren Pang, Xiaobo Xia, and Tongliang Liu.
\newblock Humanmac: Masked motion completion for human motion prediction.
\newblock In {\em ICCV}, 2023.

\bibitem{personalization}
Qiongjie Cui, Huaijiang Sun, Jianfeng Lu, Weiqing Li, Bin Li, Hongwei Yi, and Haofan Wang.
\newblock Test-time personalizable forecasting of 3d human poses.
\newblock In {\em ICCV}, 2023.

\bibitem{dozens}
Istv{\'{a}}n S{\'{a}}r{\'{a}}ndi, Alexander Hermans, and Bastian Leibe.
\newblock Learning 3d human pose estimation from dozens of datasets using a geometry-aware autoencoder to bridge between skeleton formats.
\newblock In {\em WACV}, 2023.

\bibitem{motionbert}
Wentao Zhu, Xiaoxuan Ma, Zhaoyang Liu, Libin Liu, Wayne Wu, and Yizhou Wang.
\newblock Motionbert: {A} unified perspective on learning human motion representations.
\newblock In {\em ICCV}, 2023.

\bibitem{2dposecontext}
Qitao Zhao, Ce~Zheng, Mengyuan Liu, and Chen Chen.
\newblock A single 2d pose with context is worth hundreds for 3d human pose estimation.
\newblock In {\em NeurIPS}, 2023.

\bibitem{poseformer}
Jianbin Jiao, Xina Cheng, Weijie Chen, Xiaoting Yin, Hao Shi, and Kailun Yang.
\newblock Towards precise 3d human pose estimation with multi-perspective spatial-temporal relational transformers.
\newblock In {\em IJCNN}, 2024.

\bibitem{poseformerv2}
Qitao Zhao, Ce~Zheng, Mengyuan Liu, Pichao Wang, and Chen Chen.
\newblock Poseformerv2: Exploring frequency domain for efficient and robust 3d human pose estimation.
\newblock In {\em CVPR}, 2023.

\bibitem{SMPL}
Matthew Loper, Naureen Mahmood, Javier Romero, Gerard Pons{-}Moll, and Michael~J. Black.
\newblock {SMPL:} a skinned multi-person linear model.
\newblock {\em ToG}, 34(6), 2015.

\bibitem{flowchain}
Takahiro Maeda and Norimichi Ukita.
\newblock Fast inference and update of probabilistic density estimation on trajectory prediction.
\newblock In {\em ICCV}, 2023.

\bibitem{physics}
Hiromu Taketsugu, Takeru Oba, Takahiro Maeda, Shohei Nobuhara, and Norimichi Ukita.
\newblock Physical plausibility-aware trajectory prediction via locomotion embodiment.
\newblock In {\em CVPR}, 2025.

\bibitem{MotionAug}
Takahiro Maeda and Norimichi Ukita.
\newblock Motionaug: Augmentation with physical correction for human motion prediction.
\newblock In {\em CVPR}, 2022.

\bibitem{VAE}
Diederik~P. Kingma and Max Welling.
\newblock Auto-encoding variational bayes.
\newblock In {\em ICLR}, 2014.

\bibitem{h36m}
Catalin Ionescu, Dragos Papava, Vlad Olaru, and Cristian Sminchisescu.
\newblock Human3.6m: Large scale datasets and predictive methods for 3d human sensing in natural environments.
\newblock {\em T-PAMI}, 36(7), 2014.

\bibitem{vicon}
Vicon Motion~Systems Ltd.
\newblock Vicon motion capture system.
\newblock \url{https://www.vicon.com/motion-capture}.

\bibitem{AlphaPose}
Haoshu Fang, Jiefeng Li, Hongyang Tang, Chao Xu, Haoyi Zhu, Yuliang Xiu, Yong{-}Lu Li, and Cewu Lu.
\newblock Alphapose: Whole-body regional multi-person pose estimation and tracking in real-time.
\newblock {\em T-PAMI}, 45(6), 2023.

\bibitem{BeLFusion}
Germ{\'{a}}n Barquero, Sergio Escalera, and Cristina Palmero.
\newblock Belfusion: Latent diffusion for behavior-driven human motion prediction.
\newblock In {\em ICCV}, 2023.

\bibitem{stream}
Jo{\~{a}}o Carreira, Michael King, Viorica Patraucean, Dilara Gokay, Catalin Ionescu, Yi~Yang, Daniel Zoran, Joseph Heyward, Carl Doersch, Yusuf Aytar, Dima Damen, and Andrew Zisserman.
\newblock Learning from one continuous video stream.
\newblock In {\em CVPR}, 2024.

\end{thebibliography}
\end{document}